\documentclass[letterpaper, 10 pt, conference]{ieeeconf}
\IEEEoverridecommandlockouts
\usepackage{graphicx}
\usepackage[utf8]{inputenc}
\usepackage{xcolor}
\usepackage{listings}
\usepackage{hyperref}
\definecolor{codegreen}{rgb}{0,0.6,0}
\definecolor{codegray}{rgb}{0.5,0.5,0.5}
\definecolor{codepurple}{rgb}{0.58,0,0.82}
\definecolor{backcolour}{rgb}{0.95,0.95,0.92}

\lstdefinestyle{mystyle}{
    commentstyle=\color{codegreen},
    keywordstyle=\color{magenta},
    numberstyle=\tiny\color{codegray},
    stringstyle=\color{codepurple},
    basicstyle=\ttfamily\footnotesize,
    breakatwhitespace=false,         
    breaklines=true,                 
    captionpos=b,                    
    keepspaces=true,                 
    numbers=left,                    
    numbersep=5pt,                  
    showspaces=false,                
    showstringspaces=false,
    showtabs=false,                  
    tabsize=2,
    language=Python,
}
\begin{document}

\title{\LARGE\bf VLAgents: A Policy Server for Efficient VLA Inference}

\author{Tobias Jülg$^{1}$, Khaled Gamal$^{1}$, Nisarga Nilavadi$^{1}$, Pierre Krack$^{1}$, Seongjin Bien$^{1}$,\\ Michael Krawez$^{1}$, Florian Walter$^{1,2}$ and Wolfram Burgard$^{1}$\\%
    \small{$^1$University of Technology Nuremberg $^2$Technical University of Munich}
}

\maketitle
\thispagestyle{empty}
\pagestyle{empty}

\begin{abstract}
The rapid emergence of Vision-Language-Action models (VLAs) has a significant impact on robotics. However, their deployment remains complex due to the fragmented interfaces and the inherent communication latency in distributed setups.
To address this, we introduce VLAgents, a modular policy server that abstracts VLA inferencing behind a unified Gymnasium-style protocol.
Crucially, its communication layer transparently adapts to the context by
supporting both zero-copy shared memory for high-speed simulation and compressed streaming for remote hardware.
In this work, we present the architecture of VLAgents and validate it by integrating seven policies---including OpenVLA and $\pi_0$. In a benchmark with both local and remote communication, we further demonstrate how it outperforms the default policy servers provided by OpenVLA, OpenPi, and LeRobot.
VLAgents is available at \href{https://github.com/RobotControlStack/vlagents}{github.com/RobotControlStack/vlagents}.
\end{abstract}

\section{Introduction}
The model landscape for open-source robotics foundation models is becoming increasingly rich.
Many models come with their own custom interfaces.
Evaluating, benchmarking, or extending a model, therefore, often entails writing custom deployment code to connect it to other systems and frameworks.
This makes a unified model interface highly desirable.
Yet, only a few open source solutions exist so far, most prominently LeRobot's Asynchronous Inference~\cite{lerobot}, which already has several models integrated behind a common policy server.
However, it only defines a loose dictionary-based communication interface.

In addition to code complexity, there is also a large system complexity in robotics:
AI models and robot controllers often need to run on separate physical machines, sometimes even at a remote location, as models demand more and more GPU resources.
Furthermore, as evaluation in simulation~\cite{simpler, libero, rcs} becomes increasingly popular, the robot backend should also be easily exchangeable with a simulator.
However, complex software dependencies can mutually exclude the installation of the model and simulator in the same Python environment~\cite{juelg2025refinedpolicydistillationvla}.
A policy server appears to be the natural design choice for addressing these challenges.
Many models, such as OpenVLA~\cite{openvla} and $\pi_0$~\cite{pizero}, already ship with their own policy server to facilitate the execution on a physical robot that runs on a different machine or in a different Python environment, but they are model-specific. While model-agnostic frameworks exist, they are still in their early stages of development and often lack rigid interfaces and data-aware compression.

In this work, we introduce \emph{VLAgents}, a model-agnostic Python-based inference server that provides a communication interface between models and robots.
It provides an API that is similar to Gymnasium environments~\cite{gymnasium} and is specifically tailored for Vision-Language-Action models (VLAs).
But in principle, VLAgents can be used with any type of model.
It is comprised of a flexible policy server and a corresponding client that can communicate both over the network or via efficient shared memory if executed on the same machine.
In addition, VLAgents comes with Slurm-compatible~\cite{slurm} CLI tools for checkpoint evaluations on clusters during training.

\begin{figure}[t]
	\centering
	\includegraphics[width=\linewidth]{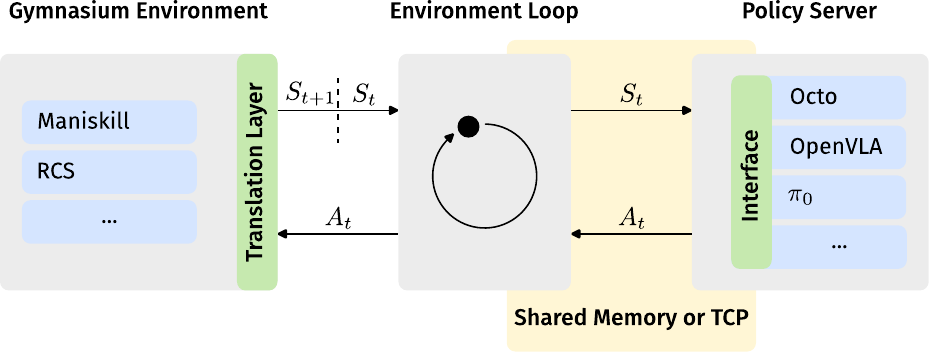}
	\caption{The architecture of VLAgents.
    Each environment implements a thin wrapper that translates observations and actions into the expected data format (see \autoref{fig:interface}).
    A central control loop takes the state and forwards it via either shared memory or a TCP connection with JPEG-encoding to the policy server.
    The policy server uses the interface from \autoref{fig:interface} to run an inference pass with the corresponding model.
    This yields an action, which is then returned and used by the environment loop to step the environment to obtain the next state.
    \vspace{-0.5cm}
    }
	\label{fig:arch}
\end{figure}

\section{Related Work}

The need for an isolated model environment has led to diverse client/server protocols.
Most common are simple synchronous raw request-response protocols.
Examples include OpenVLA~\cite{openvla}, which leverages an HTTP server implemented in FastAPI, and the OpenPi model suite~\cite{pizero, fast, pi05} with WebSocket communication.
While they work well for deploying the respective models on real robot hardware, they introduce a substantial communication overhead when used for parallel evaluation in simulation, as the payload needs to be serialized and travel through the whole network stack.
Using such model-specific servers, moreover, requires writing model-specific code for both simulation evaluation and real-world experiments.

LeRobot~\cite{lerobot} provides an asynchronous gRPC-based policy server implementation with a dictionary-based communication protocol, which is implemented for several models.
Although the framework addresses both the interface and the communication issue, it neither provides efficient shared memory-based communication nor does it perform data-aware compression.
While the approach is flexible, it requires special consideration during implementation: Since the dictionary keys are not standardized, both robots and models can require arbitrary keys for their actions and observations and considerable effort may be needed to map the keys between robots and models.
There is also no explicit protocol layer that would allow the user to define this mapping in code, which makes it difficult to apply important transformations, e.g., normalization, to the data.

\emph{VLAgents} addresses these gaps by defining a lightweight interface that transparently switches between high-performance shared memory (for simulation) and network streaming (for hardware) without code changes.
It is data-aware, allowing it to compress high-volume image data via fast JPEG encoding, while also providing the flexibility to support new data types for special use cases.

\section{Methodology}
VLAgents' architecture is driven by two main use cases:
The batched simulation evaluations on the cluster during training 
and the evaluation on physical and simulated robots for benchmarking.
Therefore, the main design objectives are to provide a unified communication interface between environments, i.e., simulations or physical robot platforms, and VLA policies, and to enable remote policy execution based on a client/server architecture.

We define a policy interface (see \autoref{fig:interface}) that wraps VLAs similar to the Gymnasium environment API.
It consists of three functions that perform model loading, resetting, and inference.
It further defines special data structures for observations and actions.
Data types required for VLAs, such as RGB input or action output, have their own dedicated typed attributes.
We also define an info dictionary that can, if required, hold any type of data.
This allows for compressing large data objects, such as images, while maintaining the flexibility to add custom data to the info dictionary.

\begin{figure}[b]
    \centering
    \begin{lstlisting}[
        language=Python,
        numbers=left,
        basicstyle=\ttfamily\footnotesize,
        numberstyle=\ttfamily\tiny,
        xleftmargin=2em,
    ]
class Obs:
    cameras: dict[str, np.ndarray] = {}
    gripper: float | None = None
    info: dict[str, Any] = {}
    
class Act:
    action: np.ndarray
    done: bool = False
    info: dict[str, Any] = {}
    
class Agent:
    def initialize(self):
        """heavy intitilzation e.g. model loading"""

    def act(self, obs: Obs) -> Act:
        """forward pass"""

    def reset(self, obs: Obs, instruction: Any, **kwargs) -> dict[str, Any]:
        """reset state, e.g. history"""\end{lstlisting}
    \caption{VLA policy interface.}
    \label{fig:interface}
\end{figure}

A policy server exposes this interface via RPyC (a TCP-based Remote Procedure Call library for Python) to a remote client.
The client is connection-aware and avoids serialization by using shared memory when running on the same host as the server.
Otherwise, RGB data are serialized using JPEG compression to reduce the data size for transport.
VLAgents can be used standalone for standard client/server communication.
Additionally, it can be used for automated evaluations as it provides an environment loop, Slurm and video recording utilities, as shown in \autoref{fig:arch}.
A wrapper layer is available to translate actions and observations into the required communication format.

\section{Results and Conclusion}
VLAgents currently integrates seven different policies, including Octo~\cite{octo}, OpenVLA~\cite{openvla}, the OpenPi suite~\cite{pizero, fast, pi05}, Diffusion Policy~\cite{dp} and V-JEPA 2~\cite{vjepa2}.
Methods that augment the input or output of the model, such as ARRO~\cite{arro}, have also been successfully evaluated.
The library comes with out-of-the-box support for Maniskill~3~\cite{maniskill} environments as well as for the Robot Control Stack (RCS)~\cite{rcs} ecosystem, which supports four different robot arms for real-world and MuJoCo-simulated experiments.
We also used VLAgents for RL-based fine-tuning of VLAs~\cite{juelg2025refinedpolicydistillationvla}, which requires batched forward passes and a low communication overhead to prevent slowing down the training speed.

\begin{figure}[t]
	\centering
	\includegraphics[width=\linewidth]{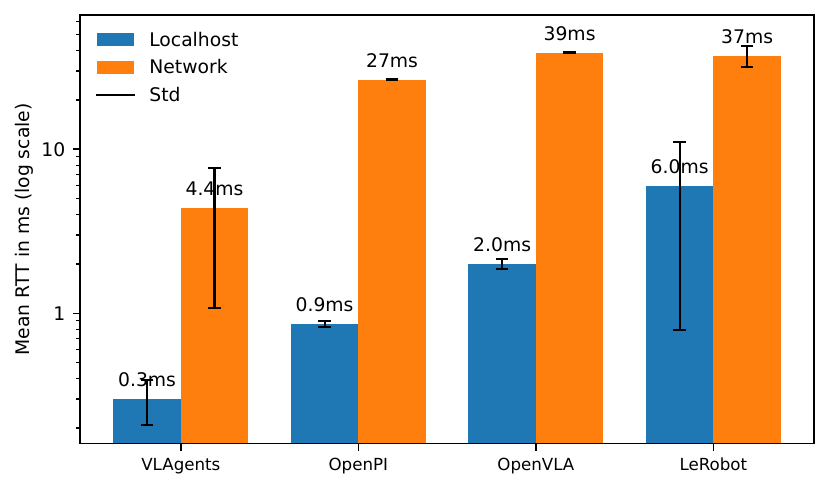}
	\caption{Mean Round-Trip Time (RTT) for different policy servers with two $224\times224$ RGB cameras.
    Localhost indicates that the client and server are running on the same machine, while network indicates execution across different machines.
    For the network setting, the machines were connected in a Local Area Network with a 1 Gbit Ethernet connection.
    \vspace{-0.5cm}
    }
	\label{fig:latency}
\end{figure}

\autoref{fig:latency} shows a comparison of plain Round-Trip Times (RTT) for client requests.
The experiment evaluates the efficiency of the serialization and the transport protocol, skipping the model's inference step on the server side.
Out of the four policy servers tested, VLAgents achieves the best performance, both in the local and the network setting.
It allows up to 220 Hz inference speed in the network deployment and introduces only 0.3 ms delay for simulated evaluations.

In conclusion, VLAgents is an efficient policy server that provides a communication interface between VLAs and robot environments, both simulated and physical, and uses data-aware compression.
Due to the usage of JPEG encoding and shared memory, VLAgents is faster than other commonly used policy servers by a factor of three.

\bibliographystyle{IEEEtran}
\bibliography{bibliography.bib}

\end{document}